\documentclass[10pt,twocolumn,letterpaper]{article}

\usepackage[pagenumbers]{iccv} 

\newcommand{\boldstart}[1]{\vspace{0.06in}\noindent{\bf #1}}

\newcommand{\tabletop}{%
    T\scalebox{0.8}{ABLE}%
    T\scalebox{0.8}{OP}\xspace%
}

\definecolor{iccvblue}{rgb}{0.21,0.49,0.74}
\usepackage[pagebackref,breaklinks,colorlinks,allcolors=iccvblue]{hyperref}

\def\paperID{12665} 
\def\confName{ICCV}
\def\confYear{2025}

\title{CObL: Toward Zero-Shot Ordinal Layering without User Prompting}

\author{Aneel Damaraju\quad Dean Hazineh\quad Todd Zickler\\
Harvard University, School of Engineering and Applied Sciences\\
{\tt\small aneeldamaraju@g.harvard.edu}
}

\begin{document}
\twocolumn[{%
\renewcommand\twocolumn[1][]{#1}%
\maketitle
\includegraphics[width=\linewidth]{figures/teaser.pdf}
\captionof{figure}{
Our model, CObL, infers a stack of occlusion-ordered \emph{object layers} that composite back to the image. (Left) We train CObL using 2250 synthetic 3D tabletop scenes created using 3D shapes with generative textures and lighting. (Right) Once trained, CObL generalizes to captured photographs of tabletops with variable numbers of novel objects. It concurrently generates a stack of amodally-completed object layers while using inference-time guidance to ensure the layers composite to the input.
\vspace{1em}}
\label{fig:teaser}
}]

\begin{abstract}
Vision benefits from grouping pixels into objects and understanding their spatial relationships, both laterally and in depth. We capture this with a scene representation comprising an occlusion-ordered stack of ``object layers,’’ each containing an isolated and amodally-completed object. To infer this representation from an image, we introduce a diffusion-based architecture named \textbf{C}oncurrent \textbf{Ob}ject \textbf{L}ayers (CObL). CObL generates a stack of object layers in parallel, using Stable Diffusion as a prior for natural objects and inference-time guidance to ensure the inferred layers composite back to the input image. We train CObL using a few thousand synthetically-generated images of multi-object tabletop scenes, and we find that it zero-shot generalizes to photographs of real-world tabletops with varying numbers of novel objects. In contrast to recent models for amodal object completion, CObL reconstructs multiple occluded objects without user prompting and without knowing the number of objects beforehand. Unlike previous models for unsupervised object-centric representation learning, CObL is not limited to the world it was trained in. 
For examples, data, and code, see our project page: \url{https://vision.seas.harvard.edu/cobl/}
\end{abstract}

\section{Introduction} \label{sec:intro}

Perceptual organization refers to the mechanisms by which pixels are structured into larger units of perceived objects and their interrelations~\cite{palmer1999vision}. Humans are very good at this, not only segmenting the visible portions of objects and inferring their figure-ground relationships, but also performing amodal completion to perceive the portions of objects that are occluded and therefore not visible. Humans can do this for a wide variety of objects, even when they are unfamiliar, partially camouflaged, and hard to name or describe.

We adopt \emph{object layers}~\cite{monnier2021layerdecomp} as a computational representation that captures the perceptual organization of an image. An object layer is a two-dimensional RGBA array (we use binary alpha channels) that conveys the location and color values of an object in isolation. When a stack of object layers is ordered according to occlusion relationships, from the background to the nearest objects, the stack composites to form an image. See~\Cref{fig:teaser}.

We suggest using this representation to explore the degree to which computer vision systems can perform zero-shot perceptual organization, where the objects or their categories have not been explicitly shown to the vision system beforehand. We capture a dataset of tabletop photographs along with their corresponding object layers by placing objects on the table one at a time, and then we measure a vision system's ability to infer the stack of object layers from the multi-object image (\cref{fig:teaser} right). To succeed at this task, a visual model must segment the visible portions of objects, infer their occlusion relationships, and infer the 2D shapes and colors of their occluded portions by amodal completion. It must do all of this without having been shown complete observations of the particular objects in the scene, without knowing the number of objects beforehand, and without any added prompting or labeling by a human.

We introduce a model for this task that leverages Stable Diffusion~\cite{rombach2022high} as a prior for natural objects. Our model uses several frozen copies of Stable Diffusion's UNet to iteratively generate a complete stack of output object layers in parallel, conditioned on the input image. To synchronize the generated layers, we tie the UNets together with trainable cross-layer attention weights, and we use inference-time guidance that encourages the output stack to composite back to the input image. Because our model generates all of the layers at the same time, we call it {\bf C}oncurrent {\bf Ob}ject {\bf L}ayers, or CObL for short.



Our model's only trainable weights are those for input conditioning and cross-layer attention. We find that we can train them to bridge the sim-to-real gap with just a few thousand synthetic tabletop scenes (\cref{fig:teaser} left), which we create using a novel generative pipeline, introduced in~\cref{sec:data}, that combines physically-accurate geometry with generated materials and lighting. Overall, we believe that CObL's performance on our dataset of tabletop photographs defines an imperfect but strong baseline for this new task.

\section{Related Work}
\label{sec:relworks}


\boldstart{Diffusion models} have been shown to generate high-quality images through iterative denoising~\cite{ho2020denoising}. We follow a trend of building on Stable Diffusion~\cite{rombach2022high}, which is a large latent diffusion model that was trained for text-to-image generation and has been effectively adapted for other tasks, including image-to-image tasks such as image editing~\cite{brooks2023instructpix2pix,zhang2023controlnet,gal2022textualinversion}, segmentation~\cite{tian2024segment}, and depth estimation~\cite{ke2024marigold}. Here, we use Stable Diffusion to help generate object layers from images.

We also follow a trend of guiding the outputs of a diffusion model using a differentiable loss that is defined on the noisy samples as they are generated (\eg, ~\cite{bartal2023multidiffusion,lee2023syncdiffusion,du2024compositional}). We introduce and use two new types of guidance losses that are complementary and effective for our task.

Our concurrent generation of multiple object layers is related to work on text-to-video generation that concurrently generates multiple video frames~\cite{ho2022videodiffusionmodels,yang2023diffusion,blattmann2023alignyourlatents,blattmann2023videoldm}. Some of these models use lateral attention between per-frame diffusion UNets~\cite{blattmann2023alignyourlatents,blattmann2023videoldm}, similar to our per-layer UNets. However, our layers have weaker relationships than the temporal correlations between video frames, and our task has much less training data. At the same time, we have the advantage of being able to use a compositing loss as guidance. 




\boldstart{Amodal completion} is the task of completing partially observed objects by inferring their occluded portions. Many previous approaches are limited to the object categories they were trained on. Some of these complete the outlines of objects without filling in their colors or textures (sometimes called amodal segmentation)~\cite{zhu2016amodalcoco,ling2020variationalamodal,ke2021BCNet,Reddy_2022_CVPR,back2021UOAIS,zhan2024amodal}, and others also complete object features such as appearance and depth~\cite{dhamo2019object,zhan2020PCNet}.  Recent works have used Stable Diffusion to achieve amodal completion for more general objects in an open-world setting~\cite{ozguroglu2024pix2gestalt, xu2023progressivemixed}, but they operate on a single object at a time, and they require human annotation to indicate which object to complete. In contrast, our model automatically completes all objects, without human prompting.




\boldstart{Inpainting} is a related task of completing arbitrary masked regions, for example using deep inpainting networks (\eg \cite{liu2018partialconv,suvorov2021lama}) or diffusion-based inpainting models (\eg \cite{lugmayr2022repaint,corneanu2024latentpaint}). These can be used for amodal completion by giving them an appropriate mask and having them complete the objects within that mask. 
Again, our model differs because it does not require an input mask.



\boldstart{Object-centric representation learning}~\cite{greff2019iodine,burgess2019monet} produces outputs that resemble object layers, but with an emphasis on being unsupervised, and in closed-world settings such as CLEVR~\cite{johnson2016clevr}, where the same objects are seen several times in different configurations. In particular, \cite{monnier2021layerdecomp} produces occlusion-ordered object layers like our model, but only for objects it has already seen repeatedly.


Recent work in this thread~\cite{seitzer2023dinosaur,akan2025SlotAdapt} has made advances toward more open-world settings by combining a slot-based object representation~\cite{locatello2020slotattention} with pre-trained features from Stable Diffusion or DINO~\cite{caron2021dino}. The output of these models is a set of slots, which are different from object layers because they are not occlusion-ordered and are only loosely associated with pixels and lack sharp boundaries. For this reason, these models' outputs are typically evaluated based on their ability to segment only the visible parts of objects, without explicit amodal completion. 




\begin{figure*}[ht!]
    \centering
    \includegraphics[width=\linewidth]{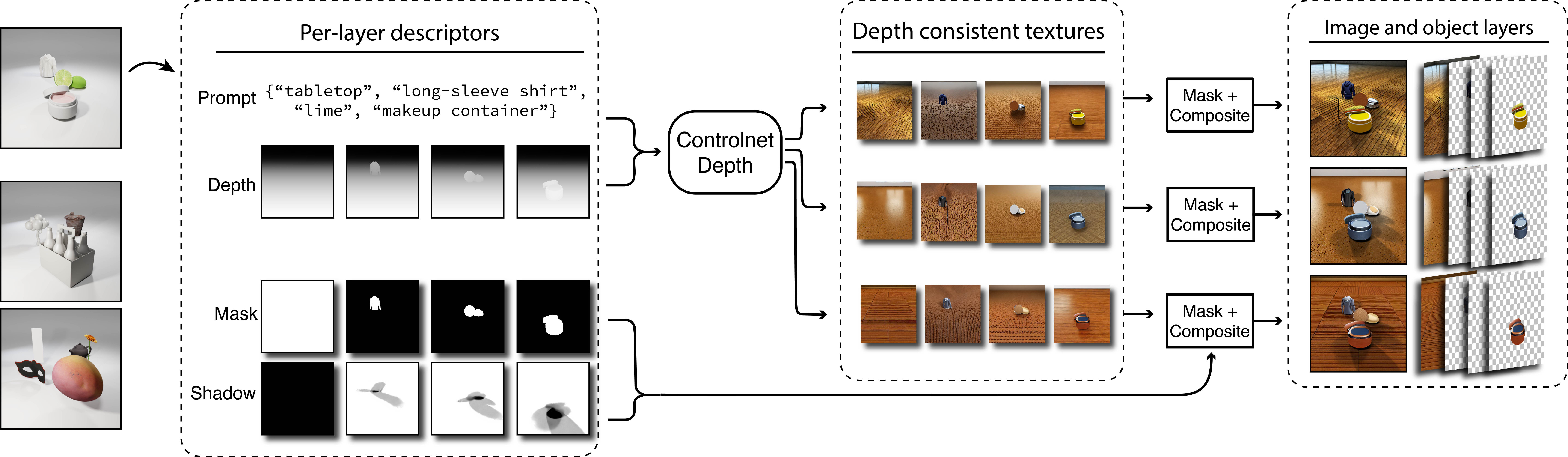}
    \caption{Our generative pipeline for synthetic training data of tabletop scenes. We use a partial rendering of randomly-placed 3D assets to create per-layer descriptors comprising depth maps, masks, shadow maps, and we combine these with object names to be used as prompts. We pass the depths and prompts to Controlnet-depth to generate multiple sets of per-layer images with different textures and reflections. Finally, we use the masks to composite the generated objects into a single image, adding shadows for more realism.}
    \label{fig:dataset}
\end{figure*}

\section{Methods}
\label{sec:methods}

A substantial challenge we face is the lack of annotated data to learn from. One can create images and object layers using rendering pipelines containing purely virtual objects (\eg CLEVR or ClevrTex~\cite{karazija2021clevrtex}) or indoor scenes (\eg~\cite{puig2018virtualhome,dhamo2019object}), but these are not general or realistic enough to bridge the sim-to-real gap. Clip Art WALT~\cite{Reddy_2022_CVPR} provides a more realistic dataset built from time-lapse images of streetscapes, but its objects are limited to cars and pedestrians. Another alternative is to manually annotate photographs, but existing datasets only include part of what we need. Some only provide amodal boundaries (\ie, amodal segmentations)~\cite{zhu2016amodalcoco,martin2001database,qi2019amodal,follmann2019learning,zhan2024amodal}, and another provides occlusion-ordering without amodal completions~\cite{lee2022instance}.

We begin by filling the need for training data, at least for tabletop scenes. We introduce a pipeline for synthesizing images and object layers using 3D modeling and text-to-image generation. We then introduce CObL and assess its ability to generalize to real images with novel objects.

\subsection{Pipeline for Synthetic Tabletop Scenes}\label{sec:data}
Our pipeline is depicted in~\Cref{fig:dataset}. It is a two-step procedure that combines (\emph{i}) partial rendering from 3D assets to create accurate geometry and shadows, and (\emph{ii}) text-to-image generation to create natural colors and textures. This two-step approach is inspired by recent work that bridges the sim-to-real gap for robotic planning~\cite{yu2024lucidsim}, and we extend their idea by using 3D object assets and rendering shadows. The resulting pipeline can efficiently create large numbers of images and object layers that, experimentally, seem to fall within Stable Diffusion's distribution of natural images. (See the synthetic validation results in \cref{fig:combined_out_all}.)

The pipeline requires a set of 3D object assets and their names or categories, which are used as prompts in the generation step. The assets do not need texture or material maps because they are not used. Also, while we focus on tabletop scenes in this paper, the idea extends to any scene geometry.

In the first step, we create virtual 3D scenes by placing random subsets of $(N-1)$ objects onto a plane, while ensuring they do not intersect but provide natural amounts of occlusion. For each scene, we render $N$ depth maps $d=(d^i)_{i=1}^N$ corresponding to each object in isolation plus an empty background, as depicted in~\cref{fig:dataset}. The depth maps are ordered by occlusion, with $d^1$ being the background and $d^N$ an unoccluded foreground object. We also render corresponding sequences of binary masks $m=(m^i)_{i=1}^N$ and shadow maps $s=(s^i)_{i=1}^N$, the latter using pre-defined lighting. This step is efficient because we do not render textures or reflections. Finally, we collect the object names into a sequence of prompt strings $p=(p^i)_{i=1}^N$. 

In the next step, we generate object layers and their corresponding image. To create the $i$th object layer $x^i$, we first generate a preliminary single-object image using a depth-conditioned text-to-image model, ControlNet-depth~\cite{zhang2023controlnet}, and the depth/prompt pair $(d^i, p^i)$; and then we mask that preliminary image using $m^i$. We then composite layers $x=(x^i)_{i=1}^N$ from back to front by recursion~\cite{wallace1981merging}:
\begin{equation}
\label{eq:compositing}
    \bar{x}^i\left((x^i)_{i=1}^N,(m^i)_{i=1}^N\right) = \frac{x^i  m^i + \bar{m}^{i-1} \bar{x}^{i-1} (1-m^i)}{\bar{m}^i + \delta},
\end{equation}
with $\bar{m}^i = \bar{m}^{i-1}  (1-m^i)  + m^i $, $\bar{m}^0$ and $\bar{x}^0$ arrays of zeros, and $\delta$ a small positive constant. This produces an initial composite, denoted by $\bar{x}^N(x,m)$, which we turn into image $I$ by adding shadows $s$ as described in the supplement.

Examples produced by our pipeline are shown in Figs.~\ref{fig:dataset} and~\ref{fig:validation_set}. Our two-step approach can generate many natural textures from the same scene geometry, which we find to be an effective form of augmentation to prevent a model from overfitting to particular shape/texture combinations.

\begin{figure*}
    \centering
    \includegraphics[width=\linewidth]{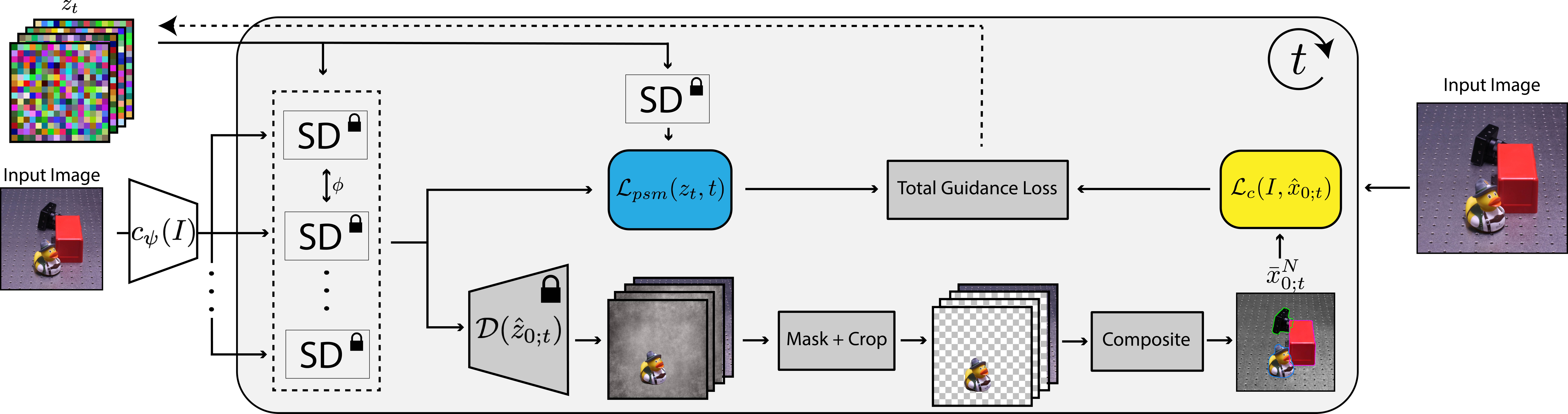}
    \caption{Our inference time iterative denoising pipeline. We compute prior score matching (blue) by comparing noise estimates of our model to those of a single pretrained Stable Diffusion UNet. We compute compositional loss (yellow) by decoding and compositing our outputs and comparing the resulting image to the input scene. }
    \label{fig:guidance}
\end{figure*}
\subsection{Model Design and Training}
Our inference task is to generate occlusion-ordered object layers $x=(x^1,\ldots, x^N)$ from an input image $I$, where $x^1$ corresponds to the background and $(x^2,\ldots,x^N)$ corresponds to the foreground object layers. Each object layer $x^i$ is an RGBA array with a binary alpha channel. For scenes containing fewer than $N-1$ objects, the unused layers should be empty, meaning their alpha values are zero.

Our model is illustrated in the left of \Cref{fig:guidance} and consists of $N$ frozen copies of the Stable Diffusion 2.1 UNet (SD). We interconnect the UNets using learnable lateral cross-attention blocks with weights $\phi$, as introduced in \cite{blattmann2023alignyourlatents}. This allows each object layer to communicate with the others while the set is concurrently generated. 

Each UNet is also conditioned on the composite image through cross-attention with the features from a learnable adapter network $c_\psi(I)$ with weights $\psi$. The adapter network first infers a pseudo-depth map using a frozen, pre-trained monocular depth estimator (MiDaS~\cite{ranftl2020midas}) and then separately produces embedding vectors for the depth map and input image following \cite{mou2023t2iadapter}. The pair are then combined with learned weights and then injected into the SD UNets which concurrently predict $N$ noise residuals, one for each object layer.

We train our parameters $(\phi,\psi)$ using pairs $(I,x)$ of images and object layers. Each object layer \(x^i\) is superimposed onto a gray canvas following \cite{xu2023progressivemixed}, converted to an RGB array, and then passed to Stable Diffusion's latent encoder $\mathcal{E}$. This yields a set of latent-encoded target layers $z=(\mathcal{E}(x_0^i))_{i=1}^N$. We then optimize the parameters $(\phi,\psi)$ attached to the frozen SD copies in an end-to-end manner, employing the standard diffusion denoising objective and noise schedule $\bar{\alpha}_t$~\cite{ho2020denoising}. Specifically, we add noise according to
\begin{equation}
\label{eq:forward_diffusion}
    z_t = \sqrt{\bar{\alpha}_t}z_{0} + \sqrt{1-\bar{\alpha}_t}\epsilon \textrm{     ,}
\end{equation}
with $\epsilon=(\epsilon^i)_{i=1}^N$ and $\epsilon^i\sim \mathcal{N}({\bf 0},{\bf I})$, and we minimize
\begin{equation}
   \mathcal{L}(\phi,\psi) = \mathbb{E}_{t,\epsilon, (I,z_0)} ||\epsilon - \epsilon_{\phi} (z_0,t, c_\psi(I))  ||^2,
\end{equation}
where $\epsilon_{\phi}(\cdot)$ are the noise predictions concurrently produced by the $N$ interconnected SD UNets.

\subsection{Guided Sampling}
We use our trained model to generate object layers for an input image $I$ as depicted in \Cref{fig:guidance}. The sampling process begins by initializing the latent layer predictions $z_{t=T}$ with random Gaussian noise. Following the standard DDIM procedure~\cite{song2022ddim}, each SD UNet predicts noise residuals that are used to estimate a single reverse diffusion step $z_{t-1}$ and an intermediate estimate of the fully-denoised object layers $\hat{z}_{0;t}$ based on the current sampling trajectory. We use guidance~\cite{lee2023syncdiffusion,bartal2023multidiffusion} to improve the transition from $z_t$ to $z_{t-1}$ by enforcing additional constraints on $\hat{z}_{0;t}$. Specifically, we update the reverse diffusion step via
\begin{align}
    \tilde{z}_t & = z_t - w \nabla_{z_t} \mathcal{L}_{g}(z_t,I) \\
    z_{t-1} & = \sqrt{\bar\alpha_{t-1}}z_{0;t} + \sqrt{1-\bar\alpha_{t-1}}\epsilon_{\phi}(\tilde{z}_t,t,c_\psi(I)),
\end{align}
where $\mathcal{L}_{g}(z_t,I)$ is a guidance loss and $w$ the step size. 

We use a loss with two complementary terms, 
\begin{equation}
    \mathcal{L}_{g} = \mathcal{L}_{c}(I, \hat{x}_{0;t}) + \lambda \mathcal{L}_{psm}(z_t,t),
\end{equation}
where $\mathcal{D}(\cdot)$ is Stable Diffusion's decoder, $\hat{x}_{0;t} = \mathcal{D}(\hat{z}_{0;t})$ are the decoded object layers, and parameter $\lambda$ controls the balance between terms. 

Loss $\mathcal{L}_{c}(I,\hat{x}_{0;t})$ is a compositional loss that measures the $\ell_2$ distance between the composite of layers $\hat{x}_{0;t}=(\hat{x}^i_{0;t})_{i=1}^N$ and input image $I$. It has the form
\begin{equation}\label{eq:comp-loss}
    \mathcal{L}_{c}(I,\hat{x}_{0;t}) = \|I-\bar{x}^N_{0;t}\|^2,
\end{equation}
where $\bar{x}^N_{0;t} = \bar{x}^N\left( \hat{x}_{0;t},\hat{m}(\hat{x}_{0;t})\right)$ is the $N$-layer compositing operation defined in~\cref{eq:compositing}, and layer masks $\hat{m}^i(\hat{x}^i_{0;t})$ are estimates obtained by separately feeding each gray-canvased layer $\hat{x}^i_{0;t}$ into a pre-trained foreground segmentation model~\cite{Qin_2020_u2net}. See the bottom row of~\Cref{fig:guidance}. At each $t$, the masks $\hat{m}^i(\hat{x}^i_{0;t})$ are created temporarily and discarded after composition.

Loss $\mathcal{L}_{psm}(z_t,t)$ encourages the latents $z_t$ to stay close to the native distribution of the original, unadapted Stable Diffusion UNet. We call this a prior score matching (PSM) loss, and it uses the same intuition as score distillation sampling~\cite{poole2022dreamfusion}. Namely, if a latent generated by our model is similar to that generated by the unadapted model, then the two models are generating latents within the same distribution. Our PSM loss is 
\begin{equation}
\mathcal{L}_{psm}(z_t,t) = \|\hat{\epsilon}_t  -  \epsilon_{\phi}(\hat{z}_{0;t}, t, c_\psi(I)) \| ^2,
\end{equation}
where $\hat{\epsilon}_t = \epsilon_\theta(\hat{z}_{0;t}, t)$ is the score estimated by the unadapted SD UNet at time $t$. We find that the PSM loss helps guide latents away from failure modes that satisfy the compositional loss but look unnatural. 

Our PSM loss is slightly different from conventional classifier-free guidance (CFG)~\cite{ho2022classifier}, and in practice we use both. CFG uses $\epsilon_\phi$ as a comparison model, while PSM uses the untied model $\epsilon_{\theta}$. This means that with CFG the stack of latents is mutually correlated, but with PSM the latent for each layer contributes separately to the loss. Intuitively, CFG promotes object layers generated from the learned multi-layer distribution, while PSM encourages each layer to be within Stable Diffusion's native distribution.

In addition, we find that including some simple non-differentiable updates to the noisy latents during sampling improves our model's ability to generalize from synthetic to real data. We periodically perform an optimal permutation of the layers to reduce the impact of bad initializations, which we find improves the average quality of generated samples. We also periodically perform automatic erase and sort operations that erase spurious hallucinations and sort empty layers. These have minimal effect on our evaluation metrics but provide cleaner-looking outputs. More details on these updates are in the supplement.

\section{Datasets}

We introduce two new datasets of tabletop scenes and their corresponding object layers. The first is a synthetic dataset generated using the pipeline from \Cref{sec:data} for training and validation. The second contains captured real-world images with objects outside of the training data distribution and is used for model evaluation.

\subsection{Synthetic training and validation data}
We create the training set following \cref{sec:data} using 600 3D object files from the Adobe Substance 3D Library~\cite{adobe_substance3d}. We use these to procedurally generate 750 tabletop configurations in Blender 4.2~\cite{blender} using the ground plane, camera and lighting positions from ClevrTex~\cite{karazija2021clevrtex}. Each configuration consists of 3 to 6 objects chosen randomly, with configurations containing fully-occluded objects being rejected. For each configuration we generate three sets of textures, producing a total of 2250 synthetic scenes. Since CObL expects a fixed number of layers for every scene, we pad stacks that have fewer than 6 objects with layers that have zero alpha.

We train on 2000 of the scenes in this dataset, and hold out the remaining 250 images as a validation set of hyperparameter tuning. This validation dataset contains objects not in the training set as well as unseen ControlNet textures.

\subsection{\tabletop dataset}

To evaluate our model's zero-shot generalization capabilities, we introduce \tabletop, a dataset of real-world tabletop images and their associated object layers. 
We then use \tabletop to compare CObL's in-the-wild performance to previous models for inpainting and amodal completion. Unlike CObL, the comparison models require an additional mask to be provided as input, indicating either the region to be inpainted or the object to be completed. We show that these models are outperformed by CObL on our dataset, even when they are given perfect oracle masks and when CObL is given none.

Our dataset includes 100 images of tabletops that each contain between 2 and 6 objects. As shown in the ablations (\Cref{sec:ablations}), we find that our model performance suffers for images with more than 4 objects, but we include scenes with 5 and 6 objects as a challenge for future work. Some examples are shown in the right of \cref{fig:teaser} and in the middle of \cref{fig:combined_out_all}. Additional examples are in the supplement. Associated with each image $I$, there is a ground-truth stack of occlusion-ordered object layers $x$ consisting of all individual unoccluded objects and the background. 

The dataset was captured using a stationary Canon EOS 40D camera with a tabletop viewing angle roughly the same as the one used in the synthetic generation pipeline. For each scene, we created the object layers by first capturing an image of the empty table and then capturing subsequent images while placing objects on the table one at a time, from the most-distant object to the nearest one. We then manually processed the set of images post-capture using a combination of Segment Anything~\cite{kirillov2023segmentanything} and manual refinement in Adobe Photoshop to obtain the final RGBA object layers.
The images and object layers are stored using sRGB colors and at a spatial resolution of $512\times 512$.

\section{Experiments}
\label{sec:evaluation}

For all experiments, we use $N=7$ instances of Stable Diffusion. 
We train parameters ($\phi,\psi$) with a learning rate of $10^{-4}$ for 100 epochs.  We use 10\% dropout of the conditioning $c_\psi(I)$ for classifier-free guidance. Training takes about 1 day on a single H100 GPU, with two scenes per batch. During inference, we use 30 DDIM steps on a linear schedule, with guidance parameters $w=10^4$ and $\lambda = 10^{-7}$.

We quantify CObL’s performance in two distinct ways. We directly measure its \emph{amodal completion} performance by assessing the shape and appearance qualities of its output object layers. We also measure the quality of the \emph{visible segmentation} that is implied by keeping only the visible portion of each object layer. This visible segmentation can be compared to other methods for panoptic segmentation.

\begin{figure*}
    \centering
    \includegraphics[width=\linewidth]{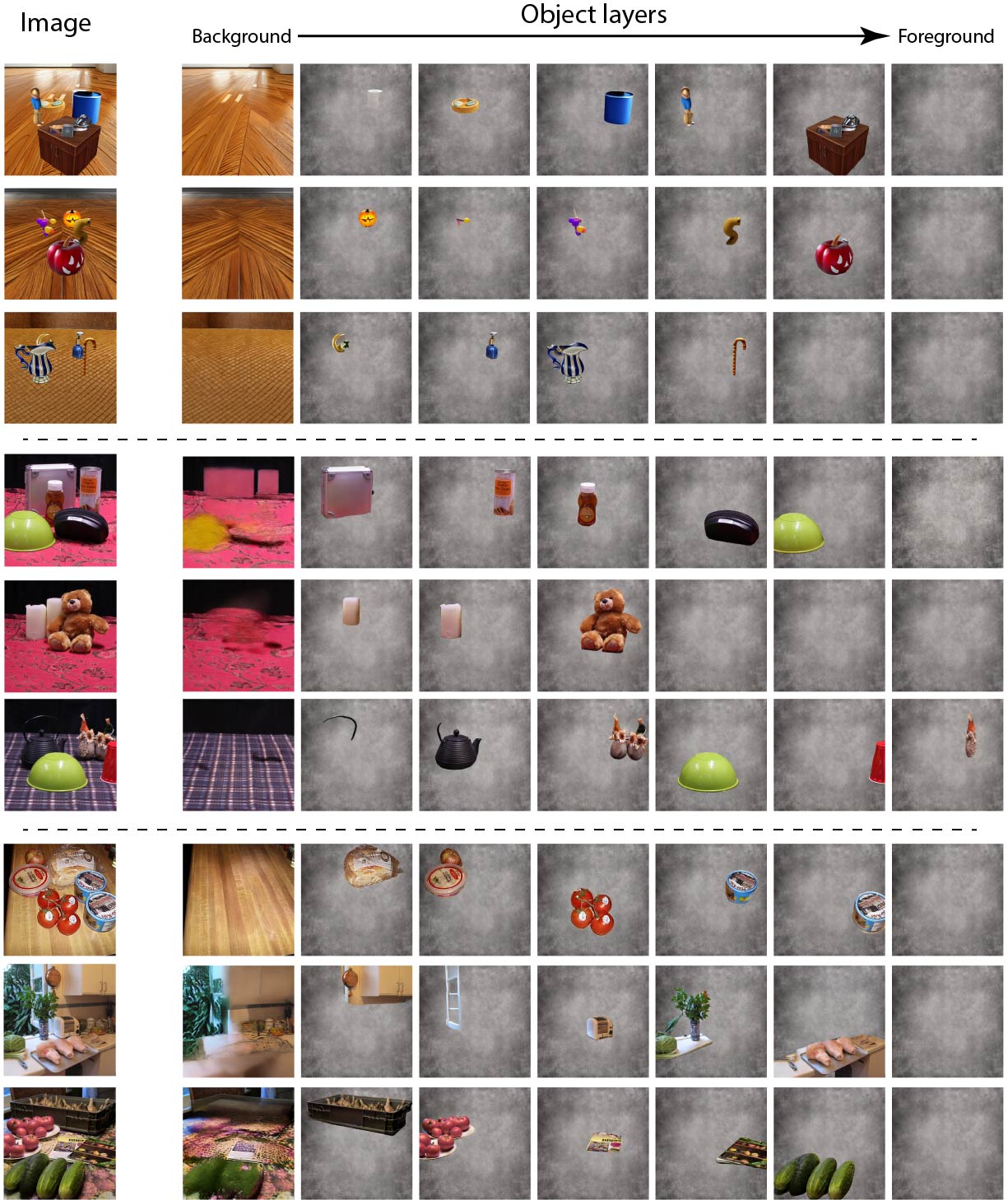}
    \caption{
CObL decomposes an image into an occlusion-ordered stack of object layers, each containing one amodally-completed object, shown here on gray backgrounds. 
(Top) Validation images within the training data distribution, and the model's corresponding output object layers. (Middle) Trained only on synthetic data, CObL performs zero-shot generalization to unseen objects in \tabletop, our real-world evaluation dataset. (Bottom) It also generalizes to some other images, such as cluttered kitchen counter-tops from~\cite{hsiao2012occlusion}, which contain new objects and viewing angles.}
    \label{fig:combined_out_all}
\end{figure*}

\begin{figure*}[ht!]
    \centering
    \includegraphics[width=\linewidth]{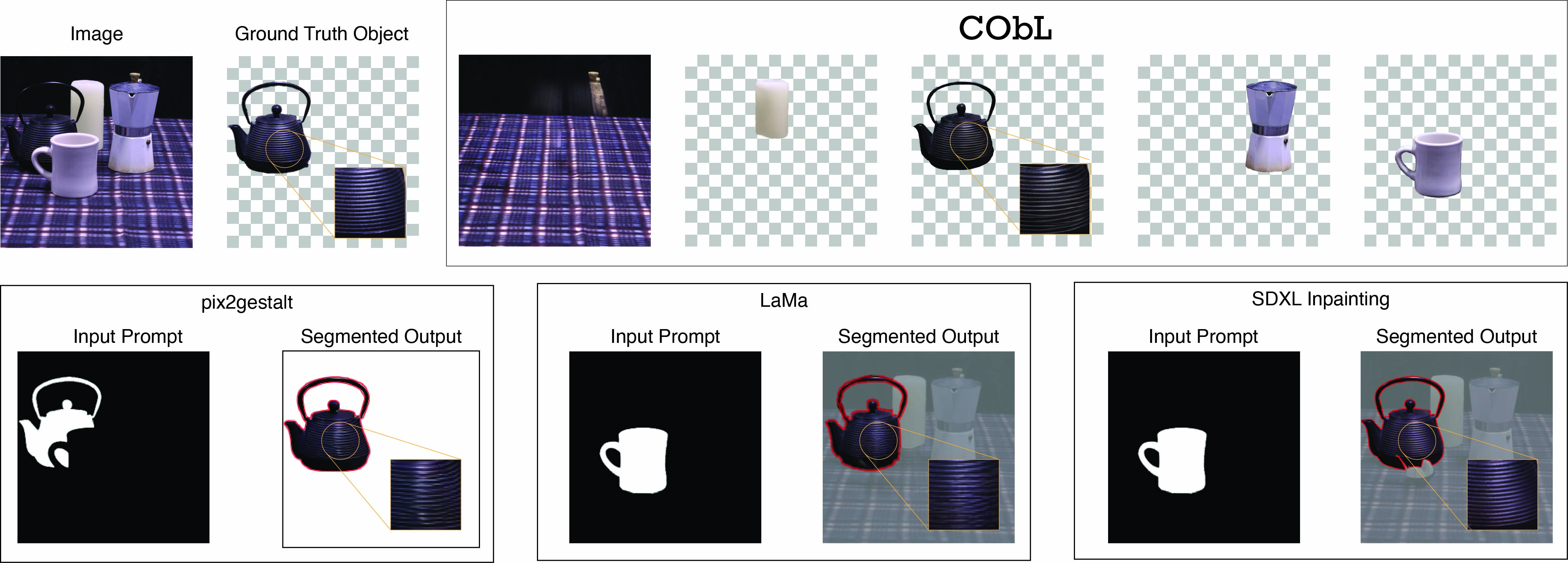}
    \caption{Comparison to previous models for amodal completion and inpainting. CObL produces all object layers without user prompting, whereas  comparison models only complete a single object, and they receive an oracle-provided mask indicating either the object to be completed (pix2gestalt) or the occluding object(s) to be removed (LaMa and SDXL Inpainting). CObL's use of compositional guidance helps it produce accurate details, such as the color and ridges of the teapot.}
    \label{fig:comparison}
\end{figure*}

\begin{table}
    \centering
    \begingroup
    \small
    \begin{tabular}{l|cc}
         &  LPIPS \phantom{$\downarrow$} & CLIP \phantom{$\downarrow$} \\
         Model and Type & (Best/Avg) $\downarrow$ & (Best/Avg) $\uparrow$ \\
        \hline
        Inpainting-GAN$^*$~\cite{suvorov2021lama} &  .113 / \phantom{.9}-\phantom{9}  & .914 / \phantom{.9}-\phantom{9} \\
        Inpainting-diffusion$^*$~\cite{podell2023sdxl}  & .373 / .384  & .760 / .758 \\
        Completion-diffusion$^*$~\cite{ozguroglu2024pix2gestalt} & .128 / .153& .889 / .872 \\
        \hline
        Object Layers (Ours) & \textbf{.094} / \textbf{.122} & \textbf{.935} / \textbf{.914} \\
    \end{tabular}
    \endgroup
    \caption{Quantitative comparison to previous inpainting and amodal completion models. For diffusion-based approaches we report best and average performance over four runs. Models with asterisks receive additional, oracle-provided masks as input.}    \label{tab:comparison}
\end{table}

\subsection{Amodal Completion}

We compare CObL's object layers to those produced by prior models for amodal completion and inpainting. This comparison puts our model at a substantial disadvantage: CObL outputs all layers without any prompting, whereas the comparison models output one layer at a time, and they receive an additional, oracle-provided input mask that is specific to each layer (see \cref{fig:comparison}). This means they have substantially more information about the input scene. 

To quantify amodal completion, we compare the estimated layers produced by our model to the ground truth object layers. We compare a true layer $x^i$ to an estimated layer $\hat{x}^j$ using both LPIPS~\cite{zhang2018lpips} and cosine similarity of their CLIP embeddings~\cite{radford2021clip}. These two scores assess visual difference and representational similarity. To obtain a single pair of scores for an entire stack $\hat{x}$, we perform Hungarian matching~\cite{kuhn1955hungarian} using LPIPS values to find the best permutation of its layers relative to $x$; and then we sum the CLIP and LPIPS scores over the $N$ per-layer scores of the permuted stack. The Hungarian matching step is important because occlusion ordering is ambiguous when objects do not overlap (see the third row of \cref{fig:combined_out_all} and \cref{fig:comp_invariance}). 

For each comparison model, we generate object layers by applying the model sequentially to remove one object at a time. See the bottom row of \cref{fig:comparison}. For the amodal completion model, diffusion-based pix2gestalt~\cite{ozguroglu2024pix2gestalt}, we provide the oracle mask that corresponds to the visible portion of the object to be completed. For the inpainting models, diffusion-based SDXL Inpainting~\cite{podell2023sdxl} and GAN-based LaMa~\cite{suvorov2021lama}, we provide the oracle mask of the occluded portion of the object to be completed.  In both cases, we segment out the model's completed object, 
paste it onto a black background, and compute LPIPS and CLIP scores as described previously.

We evaluate the comparison models and ours on a subset of \tabletop that contains scenes with 2 to 4 objects, for a total of 60 scenes. Results are in \cref{tab:comparison} and \cref{fig:comparison}.
For all diffusion-based methods, including ours, we make four runs with different noise seeds and report both the best (\ie top-1) and average scores. See \cref{sec:non-convexity} for a discussion. We find that CObL outperforms the comparisons in all amodal completion metrics, despite being asked to complete all of the layers at once, and despite being denied the advantage of having any additional oracle masks as input.


\subsection{Visible Segmentation}

To assess visible segmentation, we project all layers $\hat{x}$ to a single segmentation map by tiling the visible portions of the per-layer masks according to their occlusion relationships. The resulting segmentation map is equivalent to a conventional unlabeled panoptic segmentation~\cite{kirillov2019panoptic}. We generate ground truth panoptic segmentations by applying the same tiling technique to the ground truth object layers. We report Adjusted Rand Index (ARI) between the predicted segmentation and the true one. This measures the segmentation accuracy of the visible portions of individual objects as well as the overall quality of the scene segmentation.

CoBL achieves a top-1 ARI of 83.5\%. 
As a point of reference, we find that the panoptic segmentation model Mask2Former~\cite{cheng2021mask2former} without any finetuning achieves an ARI of 66.3\% on the same task. 

\subsection{Qualitative Results}
Some of CObL's outputs are shown in \cref{fig:combined_out_all}, including for synthetic images, images from \tabletop, and some kitchen images taken from \cite{hsiao2012occlusion}.

\begin{figure}[hbt!]
    \centering\includegraphics[width=\linewidth]{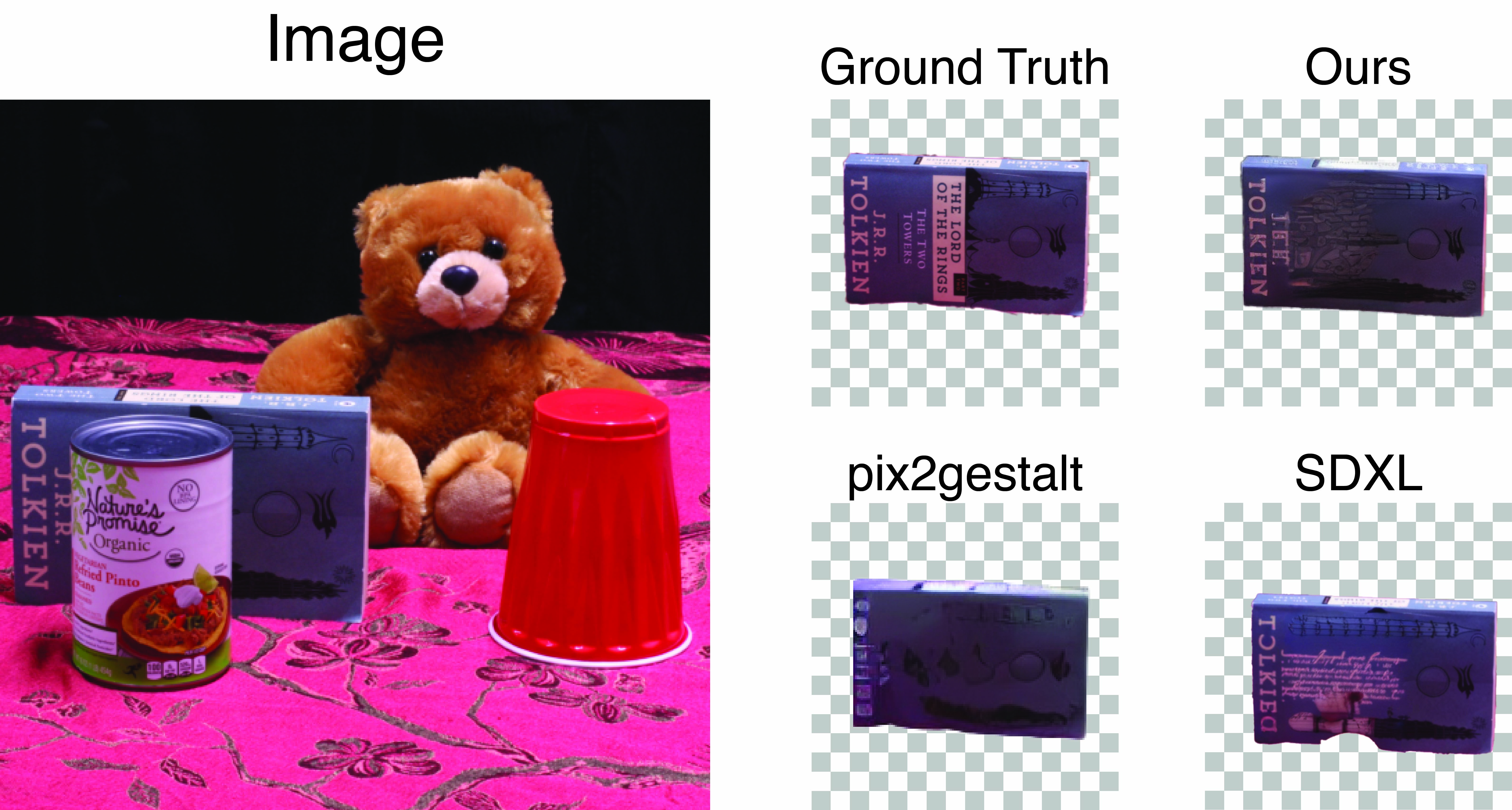}
    \caption{CObL generates faithful reconstructions of visible parts of occluded objects, such as the text on the book. Diffusion-based comparison models tend to hallucinate, even in visible regions.}
    \label{fig:comp_guidance_benefit}
\end{figure}

\Cref{fig:comparison,fig:comp_guidance_benefit} show some of CObL's outputs along with those of the comparison models. We we find that CObL does better at  preserving fine-scale details that are visible in the input image, such as the teapot in \cref{fig:comparison} and the text in \cref{fig:comp_guidance_benefit}. We attribute this to CObL's use of compositional guidance, which encourages the preservation of these details without explicitly copy-and-pasting the ground-truth visible regions as done in~\cite{lugmayr2022repaint,corneanu2024latentpaint}. 

\section{Ablations} \label{sec:ablations}
We ablate various elements of our architecture, dataset, and sampling procedure. Unless specified, we report best case LPIPS scores between CObL outputs and the \tabletop subset used in \Cref{sec:evaluation}.

\begin{figure}[t!]
    \centering
    \includegraphics[width=\linewidth]{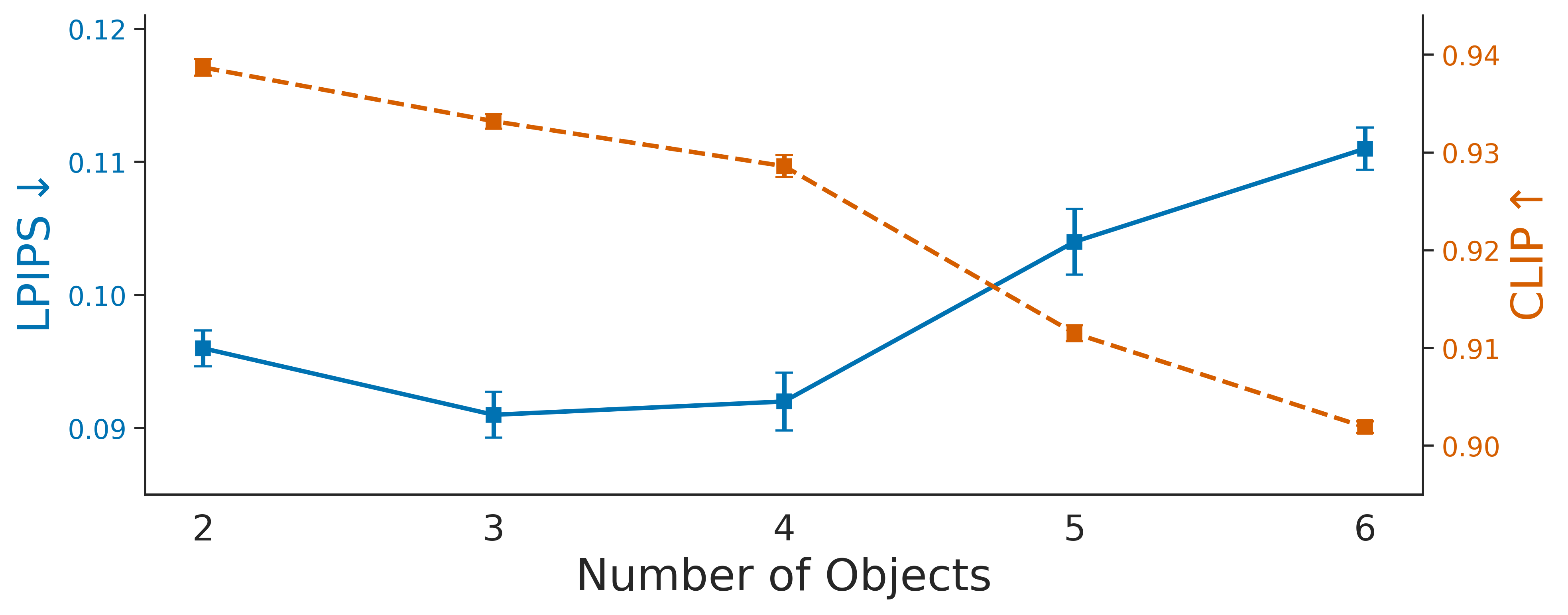}
    \caption{LPIPS and CLIP scores for varying numbers of objects in a scene. Performance degrades for more than four objects.}
    \label{fig:obj_ablation}
\end{figure}

\boldstart{Number of objects}. \Cref{fig:obj_ablation} shows output quality for increasing numbers of objects in a scene. Performance decreases for more than four objects, seemingly due to the difficulty of separating layers in increasing clutter. 

\boldstart{Depth cues}.
We train an identical version of the model but remove the MiDaS block from $c_\psi(I)$ and find LPIPS increases by $4\%$. Stable Diffusion contains a strong depth prior~\cite{ke2024marigold}, which may explain why we achieve reasonable performance without an explicit depth map. 

\boldstart{Frozen stable diffusion prior}.
Rather than using pretrained Stable Diffusion as our frozen UNet backbone, we train the model with UNet unfrozen. We report a $9\%$ increase in LPIPS on \tabletop. Without the natural image prior, out-of-distribution objects fail to be completed.

\boldstart{Guidance}.
LPIPs error increases by $8\%$ when prior score matching and compositional guidance are excluded. Image quality suffers and  layers do not composite to the scene. 

\section{Limitations and Discussion}\label{sec:limits}\label{sec:discussion}

Our formulation of guidance requires storing gradients for $N$ diffusion UNets, and this memory requirement effectively limits the maximum number of object layers in the output stack. Our model is also slower during inference compared to models that do not use inference-time guidance; and it inherits the biases of the text-to-image model that it is built upon, which in the present case leads to certain recurring failure modes associated with Stable Diffusion.

A broader limitation is that our findings and model are restricted to the world of tabletop scenes. However, within this restricted world, we discover that multiple instances of a large pretrained text-to-image model can be successfully configured to collaboratively reconstruct reasonable object-based decompositions in a zero-shot setting. A single instance of Stable Diffusion can struggle even with elemental aspects of perceptual organization, such as recognizing object parts that are disconnected by occluders~\cite{zhan2024amodal}. So our findings suggest that having interactions between multiple instances is important factor for our task. Our findings also suggest that measuring compositional quality during inference and using that to guide diffusion denoising can be effective for control of image synthesis.

\subsection*{Acknowledgments}
This work was supported in part by the NSF cooperative agreement PHY-2019786 (an NSF AI Institute, iaifi.org).







{
    \small
    \bibliographystyle{ieeenat_fullname}
    \bibliography{main}
}

\clearpage
\setcounter{page}{1}
\maketitlesupplementary

\renewcommand{\thesection}{S\arabic{section}} 
\renewcommand{\thefigure}{S\arabic{figure}}   
\renewcommand{\thetable}{S\arabic{table}}     

\setcounter{section}{0}
\setcounter{figure}{0}
\setcounter{table}{0}

\section{Details of synthetic generation pipeline}

In the Blender 3D modeling step, we place random 3D assets from our library into the scene from front to back at regular depth intervals. Each object's horizontal location is sampled randomly from all candidate locations within the frame. Object locations that lead to total object occlusion are rejected and resampled. Once placements are determined, we extract all but one object and render shadows, depth and an object mask. Creating and rendering 750 scenes takes approximately 3 hours on a NVIDIA 3080 GPU. We do not render any images using the material maps associated with the objects in our library, which greatly reduces rendering time. But for visual comparison, \cref{fig:blender_set} shows some examples of how those images would look, compared to the generated ones in \cref{fig:validation_set}.

In the texture generation step, we use the publicly available weights of pre-trained Controlnet-Depth conditioned with a rendered depth map and the prompt ``{a(n) \{object\_label\}, in a well lit and empty room, \{modifier\_string\}''. Through trial and error and with help from ChatGPT we settled on the modifier string ``hyper-realistic, photorealistic, 4k resolution, natural materials, realistic floor, realistic wall, intricate details, realistic color palette, HDR''. We run 20 steps of DDIM guidance with a CFG scale of 7.5.

To generate diverse backgrounds, we once again query ChatGPT to end up with 20 possible background prompts. Example background prompts include:
\begin{itemize}
    \item ``A cozy indoor setting showcasing detailed floor textures, \{modifier\_string\}''
    \item ``An artistic interior scene with vibrant patterns on the floor, \{modifier\_string\}''
    \item ``A luxurious indoor environment with polished wooden floors, \{modifier\_string\}''
\end{itemize}
We choose a background prompt at random, generate a scene conditioned on the input depth map, and assign it to the stack of object layers.

We then crop these outputs according to the masks to get candidate objects. Before compositing them together, we add all shadows to the background image $x_1$ by darkening pixels within the shadow regions. This update to $x_1$ is given by
\begin{equation}
    x_{1} \leftarrow x_{1}\prod_{i=1}^N (1-s_i),
\end{equation}
where we approximate the effect of shadows as a compounding decrease in background albedo. We then composite the objects as explained in the main text, see \cref{fig:validation_set} for examples of synthetic data.

\section{Architecture details}
We augment the Stable Diffusion UNet architecture with an image conditioning adapter and lateral attention blocks. We run one instance of Stable Diffusion per object layer image, tied together with lateral attention. Each UNet receives the same input text condition, using the empty string (``").

After every spatial attention layer in the UNet, we add an input adapter corresponding to image and depth cues following~\cite{mou2023t2iadapter}. The adapted cues are then passed equivalently to every concurrent UNet.  We scale the adapter weighting by a learned parameter $\alpha_{in} \in [0,1]$ such that the module is entirely skipped when $\alpha_{in} = 1$. 

Lateral attention is applied as a combination of a Conv3D and inter-layer attention block, similar to the temporal attention implementation in~\cite{blattmann2023alignyourlatents}. In our case, we extend the size of the 3D convolution kernel to $N$ as we find this helps propagate global information. Similar to the input conditioning block, each lateral attention block includes residual connection with a learned parameter $\alpha  \in [0,1]$, such that the block is skipped when $\alpha = 1$. 

We find $\alpha$ has an additional benefit beyond scaling the effect of each additional block. During inference, we can temporarily set $\alpha,\alpha_{in} =1 $ to sample denoising estimates from base Stable Diffusion. This allows us to apply prior score matching without instantiating another model.

In total, CObL has 361 million trainable parameters, about 28\% of the size of Stable Diffusion 2.1. The distribution of added model parameters across the different neural blocks are summarized in \cref{tab:added_parameters}.

\begin{table}[h]
    \label{tab:added_parameters}
    \centering
    \begin{tabular}{l r}
    \toprule
    \textbf{Network} & \textbf{Added Parameters} \\
    \midrule
    SD-UNet & 183M \\
    RGB Image Adapter  & 77.4M \\
    Depth Map Adapter & 77.0M \\
    \bottomrule
    \end{tabular}
    \caption{Parameter counts for each model component.}
\end{table}

\section{Non-differentiable guidance steps}
We employ three non-differentiable steps during inference time at equally spaced times and we find they heuristically improve output quality. We apply these updates every 5 sampling steps of inference.

\boldstart{Permute}
At inference timestep $t$, we take all predictions $(\hat{x}^1_{0;t},\ldots,\hat{x}^N_{0;t})$, exhaustively permute the ordering of these predictions and compute the resulting composition of each permutation. We then choose the permutation that minimizes the compositional loss as described in \cref{eq:comp-loss}. We find that this improves the quality of object layers of pairs of objects with very minor occlusions that may otherwise be sorted incorrectly. In practice objects with incorrect assignments will change permutations only the first time this operation is computed, but we repeat every 5 guidance steps as we find it rarely catches misclassifications in later stages of inference, with very little computational downside.

\boldstart{Erase}
As a non-differentiable update, we erase objects that are almost fully occluded. An object is considered fully occluded if less than $1\%$ of the object is visible in the composited scene. At higher scales of classifier-free guidance, the latents are less constrained by the input scene and frequently hallucinate totally unrelated objects. If a hallucination happens to be fully occluded by previous object layers, it will not be penalized by either guidance loss. This update step removes these objects by replacing the object layer with an empty layer with zero alpha. In our experiments, we do not find that object hallucinations unrelated to the object layers occur outside of these cases.

\boldstart{Sort}
To maximize output quality, we ensure the object layers are generated in a similar ordinal structure to our training data. If our model generates  layers without any objects, we note this layer as empty. We move the empty layers to the end of the layer stack so that for a scene with $k$ non-empty layers, the last $N-k$ layers will be empty. We consider a layer as empty if its alpha is nonzero for less than $0.1\%$ of the frame. 

\section{Detailed Limitations}
Beyond the high-level limitations discussed in  \cref{sec:limits}, we find failure modes in CObL that can broadly be described as ``merging" or ``splitting" failure modes. 

Merging occurs when the an object layer consists of multiple objects. We find this occurs more frequently with as the number of objects in an image increases. See the top row of \cref{fig:failure_modes}. 

Another point of failure is when an object is split across multiple layers, which generally occurs when the object has multiple visually distinct parts. In our opinion, these splits often create valid perceptual groupings. An example is shown in the bottom row of \cref{fig:failure_modes}, where the mushroom object is split across two layers along its color discontinuity.

A possible reason these failures occur is due to our use of Stable Diffusion priors, which are meant for generation of large scale scenes and not objects in particular. If we instead replaced Stable Diffusion with a model that has a stronger single image prior this issue may be resolve. However, no such model currently exists that also contains the strong prior of Stable Diffusion,.

In addition, we use an off-the-shelf mask network for foreground segmentation during inference which leads to similar issues with model bias. We find the segmentation model can lead to incorrect mask assignments for complex objects with many parts, which leads to incorrect scene compositions. A mask network specifically trained for generating object layers may improve model performance on these more complex images. 

\section{Initialization and Non-convexity} \label{sec:non-convexity}

We find that the problem of determining correct object layers is extremely non-convex. As diffusion-based generated images, our model outputs are dependent on the sample of noise used as the initial latent $z_T$. While traditional diffusion approaches can leverage this randomness to increase output diversity, poor initialization for CObL latents can result in the failure modes discussed previously. 

We find that this problem is more common in CObL than in other diffusion models because the initialization has increased total variance, as we run multiple diffusion models concurrently. However, we can leverage the exploratory nature of random initializations for improved object layer discovery. 

On many real world scenes we qualitatively find that even for the incredibly non-convex problem of occlusion-ordered object layering we can arrive at a reasonable layer stack within a handful of initializations. This motivates the ``best'' and ``average'' comparisons in \cref{sec:evaluation}. In practice, this is reflected by running CObL multiple times and choosing the most likely output.

The impact of initialization is visualized in \cref{fig:multi_trial_1}. We show the effect of using the same initialization on scenes that have similar but distinct object layer representations. Even for scenes that have different numbers of total objects, we can see the same failure modes and successes repeated when using the same initialization. We also show that we achieve different estimated object representations for the scene depending on the initialization chosen. A similar result can be seen in~\cref{fig:multi_trial_2}.

Note that regardless of if a decomposition is ``accurate" according to human perception, the object layers will generally composite back to the original scene.

\begin{figure*}
    \centering
    \includegraphics[width=1\linewidth]{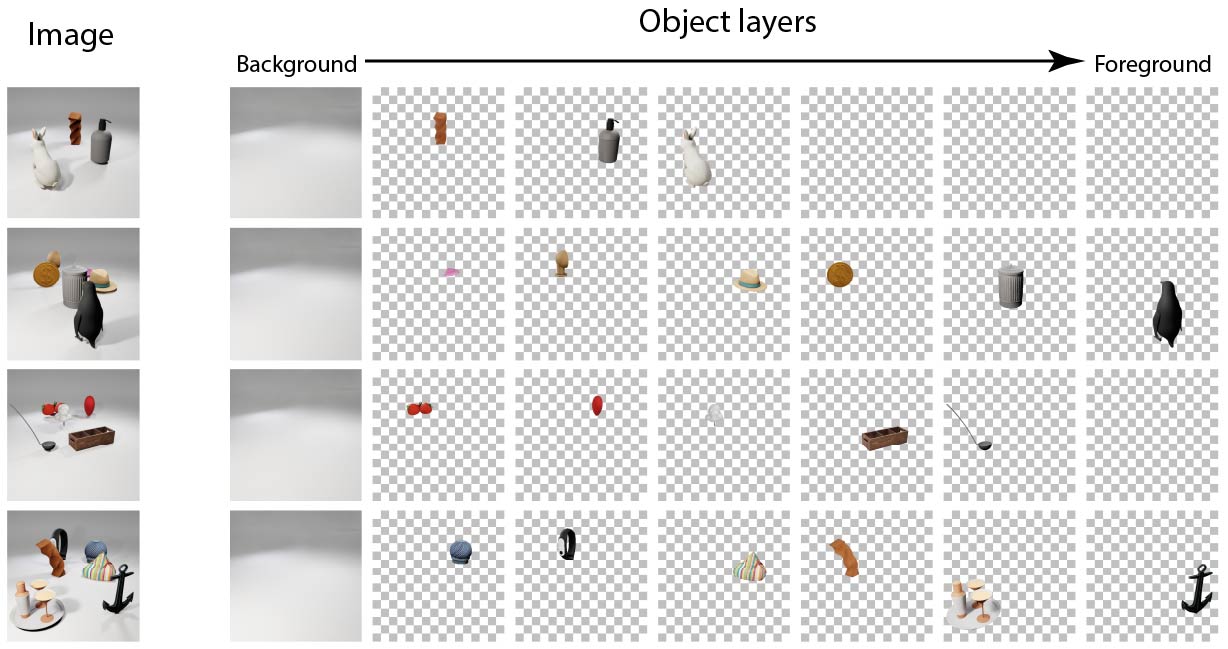}
    \caption{Renders of pre-diffusion textures. These renders are not used anywhere in the pipeline and are only shown for visualization purposes. In practice we extract the canonical object descriptors of each object for use in our pipeline, shown in \cref{fig:dataset}.}
    \label{fig:blender_set}
\end{figure*}

\begin{figure*}
    \centering
    \includegraphics[width=1\linewidth]{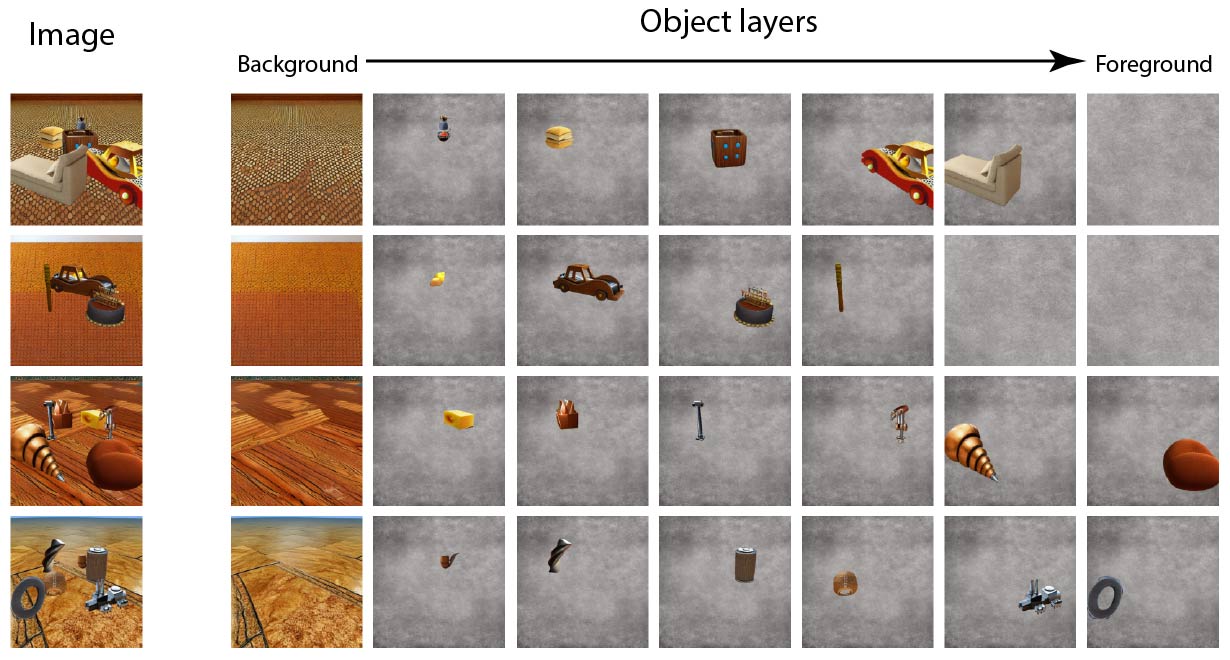}
    \caption{Random subset of our synthetic training dataset. Each object layer is placed on a gray background, and they can composite from back to front to form their corresponding image.}
    \label{fig:validation_set}
\end{figure*}

\begin{figure*}
    \centering
    \includegraphics[width=\linewidth]{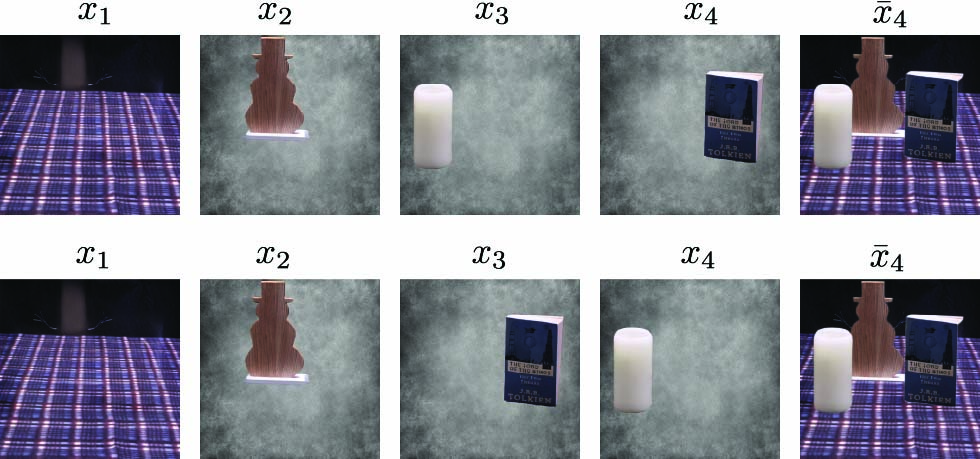}
    \caption{Two configurations of object layers that result in equivalent scene compositions. Both configurations are considered valid representations.}
    \label{fig:comp_invariance}
\end{figure*}




\begin{figure*}
    \centering
    \includegraphics[width=\linewidth]{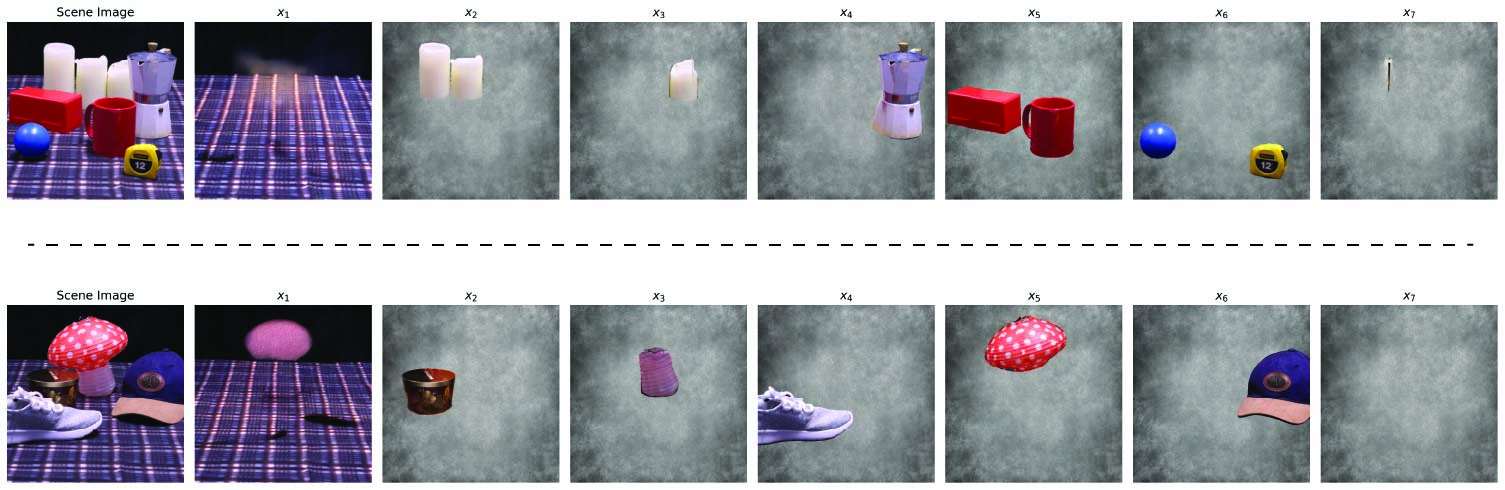}
    \caption{An illustration of the splitting and merging failure modes. (Top) A scene in which multiple disjoint objects are assigned the same layer. (Bottom) A scene in which an object (the mushroom lamp) is incorrectly characterized as two objects.}
    \label{fig:failure_modes}
\end{figure*}

\begin{figure*}
    \centering
    \includegraphics[width=\linewidth]{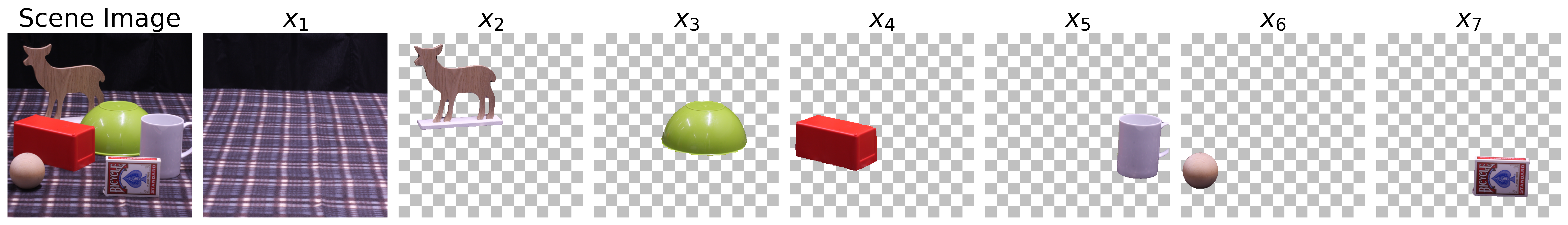}
    \caption{Ground-truth object layers for a 6 object scene in \tabletop.}
    \label{fig:tabletop_layers}
\end{figure*}

\begin{figure*}
    \centering
    \includegraphics[width=.7\linewidth]{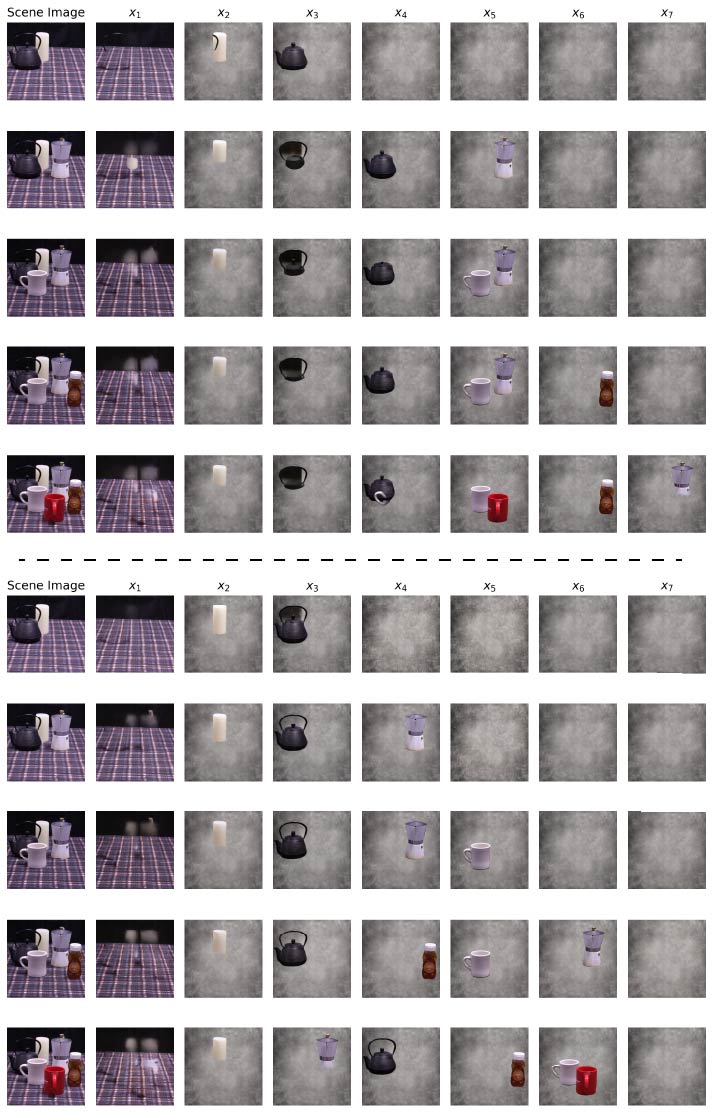}
    \caption{Learned object layers for two trials (seperated by the dashed line) of the same five scenes. In each trial, the latent initialization is kept consistent as the scenes change. Each scene varies from the previous by the addition of a single object.}
    \label{fig:multi_trial_1}
\end{figure*}

\begin{figure*}
    \centering
    \includegraphics[width=.7\linewidth]{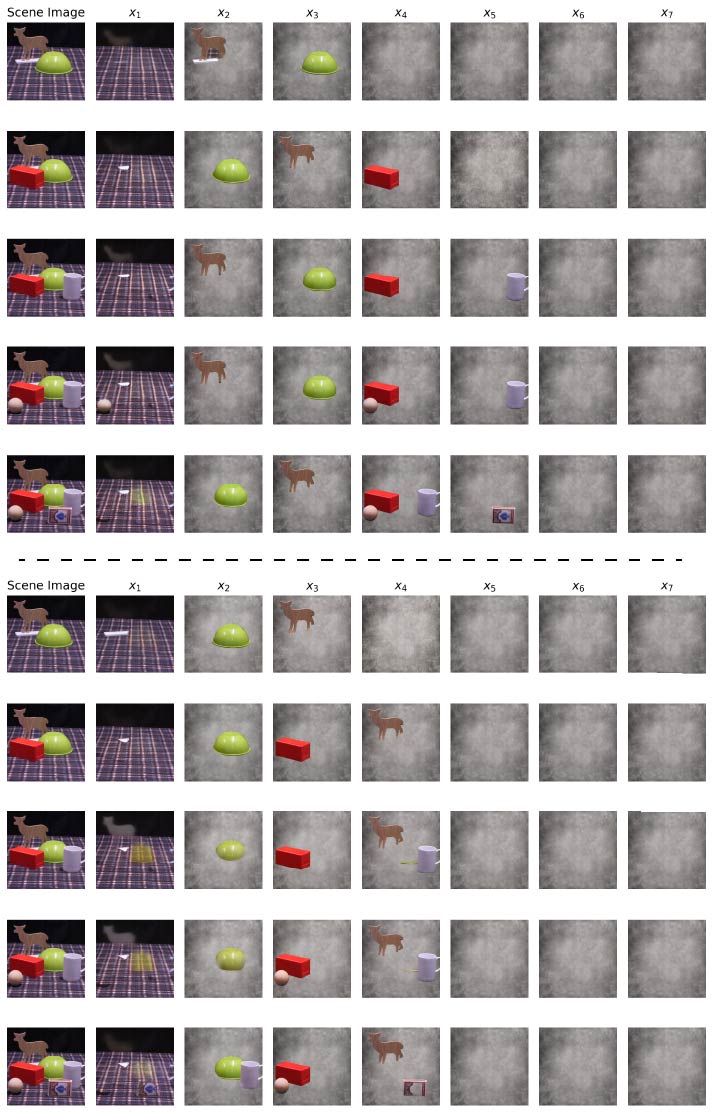}
    \caption{Learned object layers for two trials (seperated by the dashed line) of the same five scenes. In each trial, the latent initialization is kept consistent as the scenes change. Each scene varies from the previous by the addition of a single object.}
    \label{fig:multi_trial_2}
\end{figure*}

\begin{figure*}
    \centering
    \includegraphics[width=.9\linewidth]{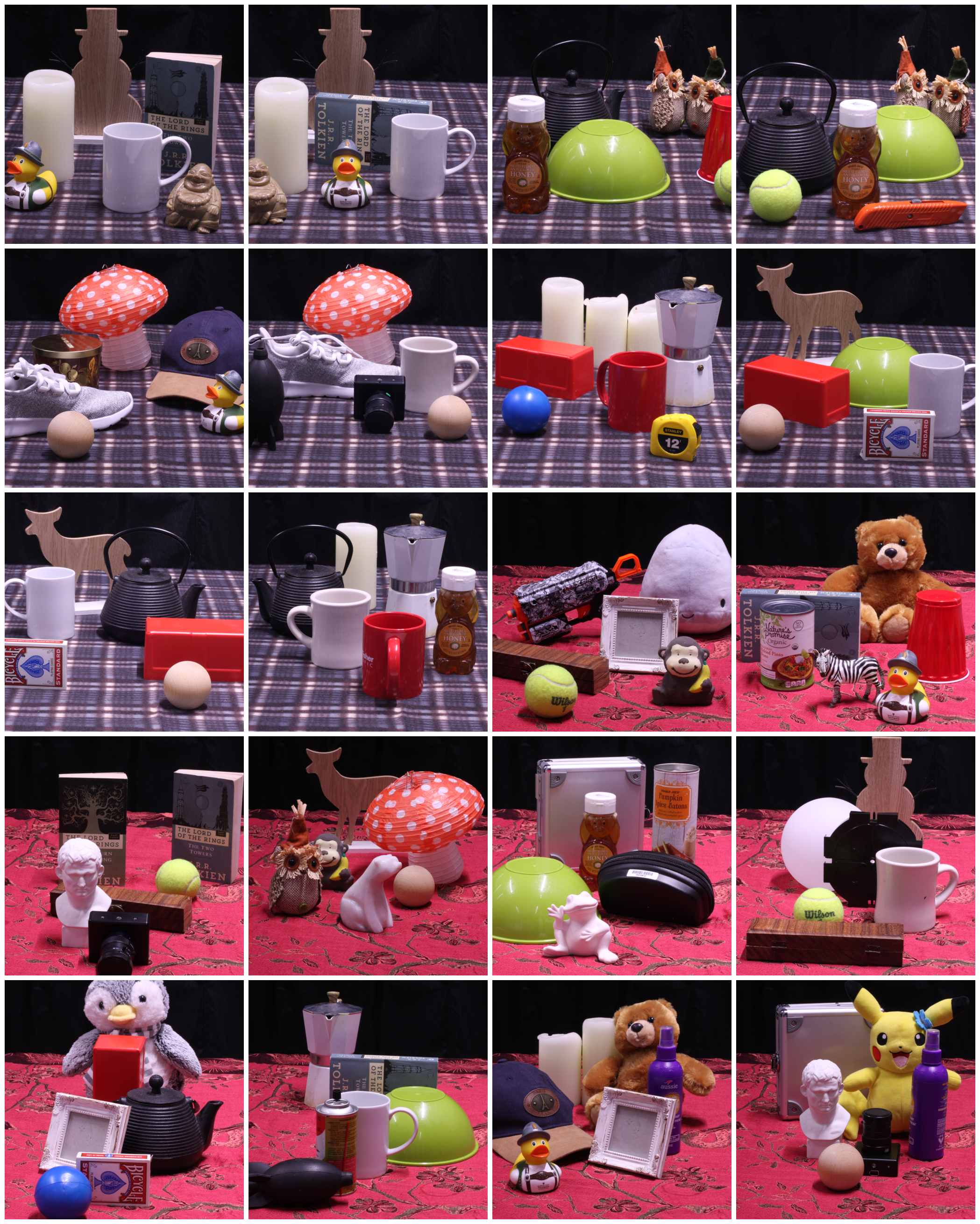}
    \caption{The set of scenes in \tabletop containing 6 objects.}
    \label{fig:tabletop_scene_set}
\end{figure*}

\end{document}